\documentclass[pdflatex,sn-mathphys-num]{sn-jnl}% Math and Physical Sciences Numbered Reference Style
\usepackage{graphicx}%
\usepackage{multirow}%
\usepackage{amsmath,amssymb,amsfonts}%
\usepackage{geometry}
\usepackage{mathrsfs}%
\usepackage[title]{appendix}%
\usepackage{xcolor}%
\usepackage{textcomp}%
\usepackage{manyfoot}%
\usepackage{booktabs}%
\usepackage{algorithm}%
\usepackage{algorithmicx}%
\usepackage{algpseudocode}%
\usepackage{listings}%
\usepackage{adjustbox}
\usepackage{booktabs}
\usepackage{longtable}

\theoremstyle{thmstyleone}%
\theoremstyle{thmstyletwo}%

\theoremstyle{thmstylethree}%

\begin{document}

\title[Article Title]{Cognitive Science-Inspired Evaluation of Core Capabilities for Object Understanding in AI}

\author*[1,3,4]{\fnm{Danaja} \sur{Rutar}}\email{rutar.danaja@gmail.com}

\author[1]{\fnm{Alva} \sur{Markelius}}\email{ajkm4@cam.ac.uk}

\author[1, 5]{\fnm{Konstantinos} \sur{Voudouris}}\email{kv301@srcf.net}

\author[1,2]{\fnm{José} \sur{Hernández-Orallo}}\email{jorallo@upv.es}

\author[1]{\fnm{Lucy} \sur{Cheke}}\email{lgc23@cam.ac.uk}

\affil[1]{\orgdiv{Leverhulme Centre for the Future of Intelligence}, \orgname{University of Cambridge}, \orgaddress{\city{Cambridge}, \country{UK}}}

\affil[2]{\orgname{Universitat Politècnica de València}, \orgaddress{\city{València},  \country{Spain}}}

\affil[3]{\orgname{Sigmund Freud University}, \orgaddress{\city{Ljubljana }, \country{Slovenia}}}

\affil[4]{\orgname{University of Primorska}, \orgaddress{\city{Koper},  \country{Slovenia}}}

\affil[5]{\orgname{Helmholtz Institute for Human-Centered AI}, \orgaddress{ \city{Neuherberg},  \country{Germany}}}

\abstract{One of the core components of our world models is ‘intuitive physics’—an understanding of objects, space, and causality. This capability enables us to predict events, plan action and navigate environments, all of which rely on a composite sense of objecthood. Despite its importance, there is no single, unified account of objecthood, though multiple theoretical frameworks provide insights. In the first part of this paper, we present a comprehensive overview of the main theoretical frameworks in objecthood research—Gestalt psychology, enactive cognition, and developmental psychology—and identify the core capabilities each framework attributes to object understanding, as well as what functional roles they play in shaping world models in biological agents. Given the foundational role of objecthood in world modelling, understanding objecthood is also essential in AI. In the second part of the paper, we evaluate how current AI paradigms approach and test objecthood capabilities compared to those in cognitive science. We define an AI paradigm as a combination of how objecthood is conceptualised, the methods used for studying objecthood, the data utilised, and the evaluation techniques. We find that, whilst benchmarks can detect that AI systems model isolated aspects of objecthood, the benchmarks cannot detect when AI systems lack functional integration across these capabilities, not solving the objecthood challenge fully. Finally, we explore novel evaluation approaches that align with the integrated vision of objecthood outlined in this paper. These methods are promising candidates for advancing from isolated object capabilities toward general-purpose AI with genuine object understanding in real-world contexts.}

\keywords{Objecthood, World Models, Core Objecthood Capabilities, Cognitive Science Paradigms, AI Evaluation}

\maketitle

\section{Introduction}

\subsection{Modelling the World}
One key premise of our existence is that we live in fundamentally open-ended and complex environments. This means that oftentimes we do not know in advance how to plan, act and what inferences to make about the events that have transpired. Despite this foregrounding challenge of human existence, adults, children and animals more generally appear to function seamlessly in their environments most of the time. Even when faced with unforeseen challenges or observing new statistical patterns we respond to them reasonably well. Consider the everyday task of going to the grocery store. When mentally planning the visit to the store, one must continuously switch between different levels of granularity. For instance, whilst it makes sense to consider specific details like the appearance of an avocado (high granularity), it makes little sense to plan each individual step required to reach the store (low granularity). Determining the appropriate level of granularity depends on both the context and the agent’s level of knowledge about the situation (e.g., making more general predictions when highly uncertain, and more specific ones when certain). Putting a glove on to place a couple of avocados in a bag and going to the scale to weigh them requires the integration and coordination of multiple sensorimotor and cognitive skills about objects. Simultaneously, one must filter out irrelevant details, such as background music or other people’s conversations, whilst directing attention to relevant information. Additionally, each trip to the store may vary depending on factors like the layout, time of day, or how crowded it is (context). These contextual changes impact navigation choices, pacing, and even the selection of substitute items when certain products are unavailable. Finally, effective grocery shopping also involves interpreting social cues and adapting to others' intentions, such as gauging when someone is about to turn into an aisle or reaching for the same product.

The ability to track and take advantage of complex environments is not limited to humans. For example, primates, dolphins and birds of the crow and parrot families have demonstrated impressive ability to spontaneously manipulate their physical environment to achieve complex goals \cite{Bastos2020, Boesch1990, Taylor2009, Visalberghi2009} even among species that did not evolve to use tools \cite{Cheke2011}. Various species have also been shown to keep track of others’ knowledge, desires and social relationships \cite{Cheney1999, Ostojic2013}. It is thus clear that humans and other animals are masters at solving every-day, multifaceted challenges.

Adaptive and context-sensitive behaviours like the one presented in the grocery shopping example in both humans and animals are underpinned by internal models of the world. These world models represent objects, space, causality, and the intentions of others, enabling agents to navigate novel circumstances with flexibility and precision \cite{lake2017building, Tenenbaum2011}. Such models allow natural agents to predict future outcomes, plan actions, and respond efficiently to unexpected challenges \cite{Shanahan2020}. World models are not merely static representations but are dynamic, continuously updated frameworks that support the agent's ability to generalise across different contexts \cite{hassabis2017neuroscience}. By using these models, natural agents can interpret new events, make decisions in real time, and solve everyday challenges seamlessly, even in unfamiliar scenarios. Through these processes, world models allow for effective adaptation and problem-solving in a wide variety of physical and social environments \cite{Bottou2014}. 

Whilst an easy task for a human, tasks described in the grocery shopping example are exactly the kinds of tasks that AI systems are still struggling with. Namely, flexibly switching between the relevant levels of granularity that monitor internal degree of uncertainty and contextual demands and making instantaneous, almost intuitive decisions. And similarly, deciding what information is relevant to attend to and what not presents a significant challenge. Possessing a robust world model allows humans to do all this by understanding the underlying concepts, rules and relationships that govern different situations in general. However, this does not imply that AI systems universally lack world models; many artificial systems do incorporate such models to varying degrees \cite{hafner2023mastering}. 

World models encompass various dimensions of understanding, including social, psychological, physical, and biological aspects of the world \cite{lake2017building}. But perhaps most fundamentally, human and animal intelligent behaviour is grounded in an implicit knowledge of objects, space, and causality—often referred to as intuitive physics \cite{lake2017building, Shanahan2020}. This tacit understanding serves as the foundation for much of our interaction with the environment, underpinning virtually every aspect of intelligent behaviour. Given its foundational role, we will focus on intuitive physics in this paper, exploring how understanding objects and their interactions shapes adaptive behaviour in both natural and artificial agents.

\subsection{Object Understanding is Central to Robust Models of the World}
As humans, we comprehend the world in terms of objects that behave in predictable ways. We use this information to predict the unfolding of future events \cite{Battaglia2013, Smith2013}, to plan actions and navigate the environment \cite{kaplan2018planning}, and to skillfully use tools \cite{Osiurak2016}. All these activities assume some understanding of what objects are, what properties they have, and how they behave - in short, these activities assume a sense of objecthood. But what exactly does this objecthood sense entail, and from what core capabilities is it built? If we were to build artificial agents with a robust sense of objecthood, what specific capabilities would we be assessing to steer their construction and test their competence?

Whilst possessing an understanding of objects is a fundamental capability of biological agents, it is far from straightforward to build artificial agents with a similar level of understanding. That is because having a true understanding of objects entails possessing various intertwined capabilities. There are likely multiple factors that could explain why artificial systems may struggle with object understanding, but we suggest that one central issue is that we lack a comprehensive theoretical account of how this understanding operates in humans and animals in the first place. Therefore, if the goal is to construct artificial agents with a robust sense of objecthood and consequently evaluate them for its presence, we should adopt a bottom-up approach of identifying most fundamental capabilities that underlie a sense of objecthood.

Various theoretical accounts of objecthood have been proposed across the cognitive sciences, underpinning independent and sometimes even opposing research agendas, making it all the harder to get a broad overview of what it means to have a robust understanding of objects (as will be discussed in the main part of the paper). To date, little attempt has been made to bring them all together. Such an overview would provide a valuable resource for AI scientists, and – we believe – represent a crucial step in making progress in building and evaluating artificial systems with a robust objecthood and intuitive physics more generally.

\subsection{Key objectives and contributions}
Objecthood is often addressed by examining isolated capabilities, such as segmentation, within a specific theoretical framework. This fragmented approach makes it challenging to define what constitutes a robust sense of objecthood more holistically—something that transcends individual capabilities. Building on this insight, the first aim of this paper is to present a comprehensive overview of the core frameworks for objecthood, drawing on Gestalt psychology, enactive cognition, and developmental psychology. In each of these frameworks, we identify core capabilities essential for understanding objects and explore the functions these capabilities serve in biological agents—an aspect especially relevant for constructing world models. Although we do not attempt to integrate these frameworks or suggest they are exhaustive, bringing them together reveals the multi-layered nature of the sense of objecthood. Also note that in this paper we only cover conceptions of object understanding for animals or humans that depend on vision, but other ways of recognising objects are possible. For example, without vision, object understanding will probably heavily depend on touch and other modalities.

Understanding these capabilities in the context of world models is critical because robust world models must encompass a nuanced understanding of objecthood. World models that integrate such capabilities enable agents to interpret dynamic environments, make reliable predictions about object behaviour, and respond flexibly to new situations. Consequently, identifying and characterising these objecthood capabilities can deepen our understanding of how both biological and artificial agents build adaptive, context-sensitive models of the world. Moreover, this characterisation is key to properly evaluate whether an AI system possesses or lacks the objecthood capabilities.

In recent years, numerous AI benchmarks related to objecthood have emerged, making it essential to assess whether they capture the complexity of object understanding seen in biological agents. This is the second aim of our paper. We examine some of the most prominent AI paradigms dealing with object understanding, comparing their methods of evaluating objecthood to cognitive science perspectives, and analysing how objecthood capabilities, as identified in the first part of the paper, are measured and evaluated in artificial systems.
From the results of the analysis of these two main objectives, our key contribution is the identification of the limitations of current AI approaches and the promising pathways to overcome those limitations. In particular, our analysis highlights the lack of functional integration across core objecthood capabilities. This fragmented focus is problematic because AI systems remain highly specialised, addressing only isolated aspects of object understanding rather than solving the objecthood challenge holistically. Capabilities related to object understanding are inherently intertwined; in biological systems, segmentation, behavioural grouping, and affordance recognition work together to create a coherent sense of objecthood. When modellers equip agents with only one capability and provide them with inputs that would normally be outputs of other object capabilities, they are doing the ‘hard (cognitive) work’ for their agents—effectively bypassing the real challenge of object understanding. This approach results in agents that excel at specific tasks, such as segmentation or affordance recognition, but lack the general, integrated object understanding involved in robust world models.

At the same time, we analyse some promising recent methods that aim to address this lack of integration. By incorporating multiple interdependent capabilities, these approaches are moving closer to a cognitively inspired model of object understanding. Embracing this integrated perspective will be critical for advancing AI systems capable of achieving a comprehensive and adaptive sense of objecthood, akin to that observed in biological agents.

\section{Core Capabilities of Object Understanding}

\textit{What is an object?} This question has received thousands of years of philosophical scrutiny, especially in the Western analytical tradition (e.g., Aristotle, Proclus, Leibniz, Descartes, Frege, Russell, Kant). It is tempting to answer this question from a metaphysical standpoint, in terms of the properties that real things in the external world have to have in order to count as an object. The problem is, there is about as much diversity in thought here as it is possible to have. The category of “object” may indeed be the most general category possible, meaning that every thing is an object, and nothing is not an object (see \cite{Rettler2017} for a review). Alternatively, objects could be collections of properties, so a particular apple is the collection of the properties red, tasty, juicy, etc. Or indeed, it could be that there are no objects at all \cite{OLearyHawthorne1995}. Some have even argued, provocatively, that there is exactly one object, the blobject, of which everything is a part \cite{horgan2000biobjectivism}.

We can sidestep these thorny metaphysical questions by instead asking the question, what is an object to an agent? Cognitive scientists have asked this question about humans and other animals. What are the perceptual units that we call objects? Take a moment and observe your surroundings. What do you see? The environment appears to be discretised into units that might correspond to, for instance, the words lamp or tree. At any moment you are also poised to act upon the world around you. And you can predict how interactions between objects will unfold in the future. Cognitive scientists behaviourally probe naive humans and animals on their intuitions about objects. As \cite{collier2023on} suggest, probing intuitive physics knowledge is concerned more with finding the norms of naive human and animal beliefs about the physical world rather than finding any metaphysical truth. In our case, the question is about what humans and animals believe about objects, and how those beliefs provide insights into the core capabilities of object understanding.

In what follows we identify three theoretical frameworks, and within them identify associated core capabilities that underlie object understanding. Drawing predominantly on Gestalt psychology, we suggest one core objecthood capability is perceptual grouping and segmentation based on static 2D, static 3D, and motion 3D cues. Drawing on developmental psychology, we propose that another core capability is grouping based on the principles of object behaviour, in short, behavioural grouping. And lastly, drawing on ecological psychology and enactive philosophy, we suggest that another core capability is identification of object affordances. We discuss each in turn. 

\subsection{Perceptual grouping and segmentation}
We are continuously presented with a stream of unstructured sensory stimuli yet our perceptual experiences appear structured. We think about the world as consisting of discrete objects that are meaningfully arranged in space. Without some process of organisation, perception would appear as a chaotic ensemble of sensations of different colours and contours as they appear on the retina, and many other cognitive activities – planning, predicting, expecting and language – would be virtually impossible, operating as they do over discrete units. A key function of perception then is to reconstruct disparate visual impressions through perceptual processes such as segmentation and grouping.

Segmentation of visual scenes and perceptual grouping (i.e., binding elements together that are disconnected at the level of the proximal stimulus \cite{Wagemans2012} are the most basic forms of abstraction that make the world manageable. Humans and animals use segmentation to abstract away from retinotopic representations of the visual field and for artificial agents segmentation represents a means of abstracting away from the arrays of pixels. The creation of objects from more primitive perceptual units makes processing more efficient because it compresses visual inputs into something more manageable before further processing, and thus reduces the units over which an organism needs to operate \cite{green2019theory, Shanahan2020}. In short, one key function of extracting “objects” via perceptual grouping and segmentation, is that of coding efficiency.

\subsubsection{Two dimensional static cues}
Systematic study of the regularities that underlie perceptual grouping was at the heart of Gestalt psychology. According to Gestalt psychology the overarching principle underpinning perceptual grouping is that of Prägnanz or “good form” which states that visual scenes are organised such that overall regularity of a visual configuration is maximised \cite{koffka1999principles, Wertheimer1982}. What spatial relations between two visual segments are good candidates to induce binding that maximises overall regularity? In his seminal work Max Wertheimer \cite{Wertheimer2017} suggested that good form could be achieved by organising visual surfaces according to certain visual properties of the sensory stimuli in the retinal image. Through a systematic manipulation of the properties of the sensory stimuli he was able to identify a set of powerful so-called Gestalt principles, which are automatically applied during perceptual grouping and result in Good form). Some of the better-known Gestalt effects are the principle of proximity (elements closer together are more likely to be grouped), the principle of symmetry (symmetrical elements are seen as a single unit), the principle of similarity (most similar elements tend to be grouped together), the principle of closure (elements that form an enclosed figure tend to be grouped together), the principle of continuity (elements aligned on a curve or a line tend to be grouped together) and the principle of parallelism (elements that are parallel with one another are likely to be grouped together) (for review see \cite{Wagemans2012}. Gestalt ideas have been often met with criticism due to their lack of empirical grounding \cite{Wagemans2012}. However, in recent years ample empirical evidence has been found in support of Gestalt principles driving segmentation in humans: e.g., the principle of closure \cite{kimchi2016perceptual, Marino2005}, the principle of connectedness \cite{Palmer1994, Scholl2001} the principle of continuity \cite{feldman2007formation, kimchi2016perceptual}, and the principle of symmetry \cite{feldman2007formation, Wagemans1997} have all been shown to play a role in perceptual grouping. No single Gestalt principle was found to be essential for perceptual grouping though, suggesting that several principles together produce impression on an object and certain combinations of Gestalt principles induce a stronger effect of “objectness” than others \cite{feldman2007formation, kimchi2016perceptual, Scholl2001}.

\subsubsection{Three dimensional static and motion cues}

Gestalt psychologists have focused solely on the role of surface 2D cues in perceptual grouping and segmentation. However, we as humans perceive the world as consisting of units that occupy space – our perception of the world is 3-D. There is a range of cues that humans use to infer 3-D objecthood. For example, humans rely on monocular cues which can be static or motion-based, and on binocular visual cues that are formed from the combined information from both eyes where the disparity between the view provided by each eye confers information about relative distance and position of the objects in the environment \cite{deangelis2000seeing}.

Static monocular depth cues, such as occlusion, relative size, relative height, texture gradient, linear perspective, aerial perspective, shading and cast shadow \cite{Bülthoff1998} are particularly useful in inferring depth and relations between objects when agents or objects around them are stationary. Static monocular depth cues are global properties of the image and therefore are more hardly inferred from small regions. For example, occlusion is harder to determine if only a small portion of an occluded object is considered. Similarly, if we only see a blue patch of a given size (on the retina), it is hard to determine whether this is large and far away (blue wall) or small and very close (blue poster), or very large and partially occluded (sky through a window). Thus, one needs to consider the overall organisation of an image to be able to determine its depth \cite{Saxena2008}.

Some monocular depth cues result from the shifts on the retinal image and are produced due to the relative movement between the agent and objects. One of the most important of these cues is motion parallax, wherein objects that are closer to an agent appear to be moving faster than objects that are farther away, due to differential changes in visual angle as the agent moves \cite{gibson1959motion, Saxena2008}. This relative motion of objects is particularly important in estimating the relative depths of surfaces \cite{Rogers1992}, and is used by many animals to estimate depth (this can be seen in the head-swaying behaviour of a cat about to pounce).

\subsubsection{Surface motions and arrangements}

Cues obtained from agent or object motion are utilised not just for judging 3-dimensionality, but also for more general object segmentation. There is some evidence to suggest that these motion cues may be used earlier in development than other Gestalt principles \cite{Spelke1990}. Three- to four-month-old infants have been shown to almost exclusively rely on motion cues in dividing visual scenes into objects, despite being sensitive to Gestalt principles in general (\cite{Spelke1993}; though see some counter evidence showing that infants of that age are also sensitive to the principle of proximity, connectedness, and common region; \cite{Bhatt2011}). In a series of experiments infants were presented with two separated, adjacent or overlapping objects combined in different ways with the aim to study how infants form object boundaries (for a review of experiments see \cite{Spelke1990}. The results unequivocally showed that three-month old infants were able to detect object boundaries solely based on the surface motions and surface arrangements (i.e., the way surfaces are organised relative to each other). Infants perceived two objects as separate entities when objects moved relative to each other even if the objects touched one another, providing evidence for the role of surface motion. Infants were also able to perceive two stationary objects as separate units when the two objects were separated in depth, providing evidence for the role of surface arrangements. Conversely, no experimental findings provided evidence that infants perceived object boundaries by forming units such that the overall regularity of a visual configuration would be maximised in accordance with the Good Form principle. For example, when stationary objects differed in textures, shapes, colours, or when they were separated in depth, three-month-olds perceived two objects as one unit although the Gestalt principle of similarity was violated. Interestingly, adults were found to detect object boundaries by relying on all three types of cues: surface motions, surface arrangements and surface characteristics \cite{kestenbaum1987perception, Spelke1993}. Similar findings were also obtained in studying how infants infer object unity \cite{Spelke1990}. Contrary to Gestalt predictions, findings from multiple studies show no evidence that when infants see partly occluded objects they group object surfaces into units that are maximally regular and simple. For example, when infants saw the ends of a partly centre-occluded object move together behind the occluder they perceived that object as one connected unit (see also \cite{gerhardstein2004detection}, for similar conclusions). In other words, infants did not perceive the partly occluded object as several objects but as one continuous object. Perception of such moving objects was also not affected by objects’ configurational properties: Infants perceived objects whose visible surfaces were not symmetric and homogeneous in texture and colour equally strongly as a single unit compared to the objects that were uniform in texture and colour \cite{Spelke1990} (thus violating the principle of similarity). Again, this is unlike adult perception where centre-occluded object identification was affected by both surface arrangements, motions, and static Gestalt properties.

What does this developmental pattern tell us about the nature of object perception? The transition from purely motion-based to motion-based coupled with Gestalt principles might be indicative of a shift from a model-free to a model-based learning \cite{Spelke1999}. Given that infants appear to be sensitive to Gestalt principles in contexts other than object perception, it might be that using Gestalt principles in the context of object perception is something that can occur only after the development of models of objects and their properties. The idea suggests that first, infants recognise moving surfaces and arrangements as relevant inputs for visual segmentation. This input enables them to build coarse grained models of objects. After having built these first, crude models of objects infants can start focusing on more detailed regularities of objects summarised by the Gestalt principles, allowing them to perform more sophisticated object models and segmentation strategies. In short, object perception might progress from global, coarse grained visual impressions built predominantly on surface movements and arrangements to local, detailed analysis of visual scenes based on static Gestalt principles. Such an account awaits empirical investigation.  

\subsection{Behavioural Grouping}
Object understanding is not limited to perceptual grouping and segmentation involved in object identification. For grouping and segmentation to be useful to an agent, they must facilitate prediction about how that object is likely to behave. This understanding of object behaviour then facilitates further sophistication in object identification, particularly in complex and dynamic environments. The above-mentioned analysis of surface arrangements and motions might in fact play a crucial role in the formation of the earliest physical expectations about object behaviour. In other words, as the principles of object behaviour are about objects in space and time, the most likely input for initial learning about these principles are surface arrangements and motions. 

For example, infants from a very young age believe that objects maintain their size and shape when moving (principle of rigidity), that they move as connected, coherent wholes (principle of cohesion) separately from one another (principle of boundedness), that objects act upon each other only when in contact (principle of no action at a distance), and that objects obey principles of persistence (objects retain their individual properties), continuity (objects continue to exist) and solidity (objects remain cohesive) (e.g., \cite{Baillargeon2002, Spelke1999}). By their first birthday, infants have acquired several other principles of object behaviour, such as inertia (objects will continue to move in a straight line unless acted upon by another force), support (objects need support to stay elevated), containment (object placed inside a container continue to exist) and collision (when one object hits another, there should be a reaction) \cite{Baillargeon2004, hespos2008young}. These early emerging principles of object behaviour, once mastered, guide subsequent learning about objects and are used as additional means of visual segmentation (ref). The process of learning to perceive objects is thus intrinsically intertwined with being able to reason, or at least make predictions, about object behaviour \cite{Spelke2013}. The mechanisms and core capabilities involved in object understanding enable agents to perceive the world beyond their immediate surroundings and allow them to understand properties of objects that cannot be sensed. Thus, the ability to perceive objects (i.e., perception) is directly related to the ability to reason about objects and their behaviour (i.e., conception) \cite{Spelke2013}.

For Spelke, the initial focus on moving surfaces is a developmental step and, gradually, object perception is the result of a combination of Gestalt principles and the analysis of the surface motions and arrangements. However, Matosa has argued that perceiving objects is exclusively based on analysis of moving surfaces that behave in predictable, object-like ways \cite{Matthen2021}. Developmental processes do not shift the focus in which features are used in object perception. When we see objects, we see causally coherent integrated wholes that perform uniform actions, are spatiotemporally extended and continuous collections of matter that occupy spatial locations and possess features that are not seen from all perspectives. Objects thus cannot be seen independent of their defining features. These features are not captured by perceptual grouping and visual segmentation, indeed the least consequential thing about objects is that they are collections of surface points grouped together by some static principles of perceptual grouping. Even more, given that perceptual grouping and segmentation do not capture any of the defining features of objects described above, perceptual grouping and segmentation are neither sufficient nor necessary aspects of object perception \cite{hassabis2017neuroscience}.

Consider the following example that showcases this idea. Imagine observing a bird flying behind a set of trees. Although you cannot continuously see the bird, your world model allows you to predict its trajectory based on an understanding of how birds typically move: steady, linear paths with occasional flaps. This predictability helps you perceive the bird as a coherent object, even when occluded. Contrast this with observing a squirrel, which moves in short, erratic jumps from tree to tree. Your world model adjusts to this different type of movement, predicting the squirrel’s likely positions based on its characteristic behaviour. A world model integrates knowledge of object-specific behaviours and uses them to make dynamic, context-sensitive predictions. These differing predictions, grounded in our intuitive understanding of the physical and behavioural properties of animals, help us identify and differentiate these objects in motion. In short, whilst movement can help us detect the presence of an object, it alone is not enough to fully perceive it. To recognise an object as distinct from its surroundings, we rely on our ability to predict its behaviour in a consistent and meaningful way, using knowledge from our internal world model.

The above account, whilst valuable, perhaps represents an extreme end of this theoretical framework in emphasising the sufficiency of motion information in inferring objects. We believe that different theoretical accounts have different explanatory targets (in what kinds of objects and perceptual processes they aim to explain). Therefore, they might be sufficient for explaining certain kinds of capabilities and certain kinds of perceptual processes involved in objecthood but not others. For example, theoretical frameworks emphasising the role of principles of object behaviour and the importance of motion in perceiving objects will be better for explaining how we perceive objects where motion plays a crucial role, for example in perceiving camouflaged animals or waves. And frameworks emphasising the role of grouping cues will be better at explaining initial stages of scene construction and how information in the early visual cortex is used to infer initial representations of objects. One could perhaps argue that the difference between different theoretical frameworks for explaining objecthood is that the framework investigating behavioural grouping focuses more on top-down aspects of objecthood (thus emphasising the role of cognitive processes) and the Gestalt psychology focuses more on the bottom-up processes involved in objecthood. 

\subsection{Identification of Affordances}

The two theoretical frameworks discussed above assume different functions of perception. The account built on Gestalt psychology assumes that perception is for the segmentation of the visual environment, that then allows for further reasoning on those "segments": You see an object and this allows you to learn how it will behave. On the other hand, the framework built on developmental psychology highlights the importance of understanding object behaviour in the process of perceiving and delineating a visual environment: You see things as an object because they behave like an object. This theoretical framework that we present next is built on yet another assumption, i.e., that perception is for, and intertwined with, action \cite{gibson2014ecological}. And the core capability that fulfils this function is identification of affordances.
Agents get to know their environments by moving around, changing the position of their heads, eyes and bodies. Thus, for embodied agents, perception is inherently an activity \cite{Anderson2017}. This is nicely summarised in the following passage: “perceptual experience acquires content thanks to our possession of bodily skills. What we perceive is determined by what we do (or what we know how to do); it is determined by what we are ready to do…we enact our perceptual experience: we act it out” [47, p. 1, italics in original]. The physical environment is rich with possibilities for (inter)action and agents are able to identify and act on these possibilities – these affordances. Affordances are action possibilities formed based on the relationship between the agent and the environment in which the agent is situated \cite{gibson2014ecological}. As information present in affordances entails information about the environment and the agent in concert, exteroception (perception of the external world) is inextricably linked with proprioception (perception of the movements of one’s own body – note that the embodiment assumption holds here). Thus, to perceive an object is, according to this account, to see its motoric value and to co-perceive one’s own interactive potential with it \cite{Chemero2003, gibson2014ecological, Warren2021}. This relational aspect is at the heart of Gibson’s theory of affordances, and it predicts that different animals with different physical and perceptual characteristics will identify different affordances in the same object, and thus perceive the object differently. For example, when I see a cup filled with liquid, I see it as affording quenching my thirst (as I am able to grasp the cup and raise it to my lips). A spider's perception of the same cup on the other hand will be entirely different – when the cup is empty it can offer a refuge from a hungry toad. But, when full, it affords drowning!
Gibson’s theory of affordances provided a good theoretical basis, however it did not specify the exact link between the identification of affordances and object perception. Tucker and Ellis \cite{Tucker1998, Tucker2001} were one of the first ones to systematically investigate this relationship. In a series of experiments they studied whether participants automatically generate motor or affordance representations when they are presented within an object even in the absence of the intent to act \cite{Tucker1998, Tucker2001}. In one experiment participants were asked to press the left or the right button to indicate whether the object image they had just seen was upright or inverted. Objects presented to the participants had a pronounced right or left-handed affordance, for example, a pan handle oriented to the left affords left-handed grip. The results of the study showed that the participants were faster to respond if the task-irrelevant horizontal orientation of the object was congruent with the required response. For instance, if participants were instructed to make right responses to upright objects and left responses to inverted objects, these were more accurate and faster, if the presented object had its graspable region to the right when upright and to the left when inverted. Building on the Tucker and Ellis \cite{Tucker1998} study Boroditsky and colleagues \cite{flusberg2010motor} conducted another study examining whether affordance representations contribute to the perception of objects. In one of their experiments  participants first saw an image of a hand and then they were presented with an ambiguous object drawing. They were asked to name that object and to identify whether the hand they had seen was left or right. The results showed that the responses were biased towards the interpretation that was congruent with the grasp type of the hand prime. For example, when participants were primed with a hand positioned into a power grasp they were more likely to interpret an ambiguous image as affording a power grasp. Based on this the researchers concluded that a motor/affordance representation  can play a causal role in object perception. After these initial studies others have found similar results suggesting that affordance representation formed upon seeing an object are constitutive of the perception of objects \cite{helbig2006role}.The two theoretical frameworks discussed above assume different functions of perception. The account built on Gestalt psychology assumes that perception is for the segmentation of the visual environment, that then allows for further reasoning on those "segments": You see an object and this allows you to learn how it will behave. On the other hand, the framework built on developmental psychology highlights the importance of understanding object behaviour in the process of perceiving and delineating a visual environment: You see things as an object because they behave like an object. This theoretical framework that we present next is built on yet another assumption, i.e., that perception is for, and intertwined with, action \cite{gibson2014ecological}. And the core capability that fulfils this function is identification of affordances.
Agents get to know their environments by moving around, changing the position of their heads, eyes and bodies. Thus, for embodied agents, perception is inherently an activity \cite{Anderson2017}. This is nicely summarised in the following passage: “perceptual experience acquires content thanks to our possession of bodily skills. What we perceive is determined by what we do (or what we know how to do); it is determined by what we are ready to do…we enact our perceptual experience: we act it out” \cite{Noe2004}, (p. 1, italics in original). The physical environment is rich with possibilities for (inter)action and agents are able to identify and act on these possibilities – these affordances. Affordances are action possibilities formed based on the relationship between the agent and the environment in which the agent is situated \cite{gibson2014ecological}. As information present in affordances entails information about the environment and the agent in concert, exteroception (perception of the external world) is inextricably linked with proprioception (perception of the movements of one’s own body – note that the embodiment assumption holds here). Thus, to perceive an object is, according to this account, to see its motoric value and to co-perceive one’s own interactive potential with it \cite{Chemero2003, gibson2014ecological, Warren2021}. This relational aspect is at the heart of Gibson’s theory of affordances, and it predicts that different animals with different physical and perceptual characteristics will identify different affordances in the same object, and thus perceive the object differently. For example, when I see a cup filled with liquid, I see it as affording quenching my thirst (as I am able to grasp the cup and raise it to my lips). A spider's perception of the same cup on the other hand will be entirely different – when the cup is empty it can offer a refuge from a hungry toad. But, when full, it affords drowning!
Gibson’s theory of affordances provided a good theoretical basis, however it did not specify the exact link between the identification of affordances and object perception. Tucker and Ellis \cite{Tucker1998, Tucker2001} were one of the first ones to systematically investigate this relationship. In a series of experiments they studied whether participants automatically generate motor or affordance representations when they are presented within an object even in the absence of the intent to act \cite{Tucker1998, Tucker2001}. In one experiment participants were asked to press the left or the right button to indicate whether the object image they had just seen was upright or inverted. Objects presented to the participants had a pronounced right or left-handed affordance, for example, a pan handle oriented to the left affords left-handed grip. The results of the study showed that the participants were faster to respond if the task-irrelevant horizontal orientation of the object was congruent with the required response. For instance, if participants were instructed to make right responses to upright objects and left responses to inverted objects, these were more accurate and faster, if the presented object had its graspable region to the right when upright and to the left when inverted. Building on the Tucker and Ellis \cite{Tucker1998} study Boroditsky and colleagues \cite{flusberg2010motor} conducted another study examining whether affordance representations contribute to the perception of objects. In one of their experiments  participants first saw an image of a hand and then they were presented with an ambiguous object drawing. They were asked to name that object and to identify whether the hand they had seen was left or right. The results showed that the responses were biased towards the interpretation that was congruent with the grasp type of the hand prime. For example, when participants were primed with a hand positioned into a power grasp they were more likely to interpret an ambiguous image as affording a power grasp. Based on this the researchers concluded that a motor/affordance representation  can play a causal role in object perception. After these initial studies others have found similar results suggesting that affordance representation formed upon seeing an object are constitutive of the perception of objects  [e.g., 53].

\subsection{Interaction between capabilities}

Above, we presented three core capabilities involved in object understanding as if they were independent of each other and hence did not discuss how they might be intertwined. This is partly because doing so would go beyond the scope of this paper, but another reason is that this is how objecthood has traditionally been studied. Different communities of scientists focus on different aspects of what it means to understand objects, and our division is a reflection of that.

Nevertheless, we want to emphasise that the core capabilities presented here are, in fact, inherently intertwined in that they likely specify inputs for each other and hence feed into one another. For example, to understand abstract principles of object behaviour, it is necessary to possess some notion of a segmented surface to which this behaviour refers. Alternatively, it seems insufficient to think about objects only in terms of segmented units that do not also afford action. So, whilst cognitive scientists might independently study various capabilities involved in object understanding, human object understanding is much more deeply integrated across the various capabilities presented here. The ramifications of this segregated treatment of object understanding will be discussed more in depth next, in the context of AI, where the lack of crosstalk presents an even bigger issue.

\section{Assessing Objecthood Understanding in AI}

In this section, we evaluate the extent to which current AI paradigms have captured the capabilities associated with object understanding, as presented in the first part of the paper. A paradigm is a conceptualisation of objecthood, methods that emanate from it, data that represents the main features (demands) in the conceptualisation and methods of evaluation for that conceptualisation and data. We identify the core AI paradigms used to test these proposed capabilities and then analyse how they have implemented and evaluated these core objecthood capabilities. Note that the mapping between the capabilities and the AI paradigms discussed is not direct in that some paradigms integrate more than one cognitive capability and some cognitive capabilities appear across several AI paradigms. The different paradigms in which objecthood sense has been tested in AI are not viewed as distinct categories but as points on a conceptual plane, as illustrated in Fig. 1. This plane spans two dimensions: Knowledge about the world and Interaction with the world. These dimensions reflect the extent to which each approach either builds on pre-existing world knowledge or interacts with the environment to gather information. A full classification of approaches would be based on varying criteria for how each method conceptualises and processes object information, whether through visual features, learned representations, or interactive approaches. 

The Bottom-left quadrant (3) represents methodologies where the system processes visual input to identify and segment objects, relying on computer vision and geometrical features for knowledge. This approach involves minimal real-time interaction with the world but focuses on extracting object information from pre-existing visual datasets \cite{Chai2021, gao2023deep}. This is related, but not equal to the top level in Fig. 1. The Top-left quadrant (1) illustrates approaches that use, for example, generative AI such as large language models (LLMs) to generate or infer contextual object knowledge based on combining, e.g., visual and textual data \cite{huang2024combining, kirillov2023segment, Li2024, Wang2023}. This vaguely captures elements that we see in the other two levels in Fig. 1, but the exact relation will be revisited at the end of the section. For instance, this approach relies heavily on pre-existing knowledge about the world but involves minimal interaction with the world for real-time data acquisition. Deep Reinforcement Learning (DRL) agents and game engine physics (the Bottom-right quadrant, 4) are methods where agents in interaction with the world often through game engines or simulated environments learn object understanding through repeated engagement with the environment \cite{Voudouris2024}, using physics-based game engines to refine physical reasoning about objects \cite{Wu2015}. The Top-right quadrant (2) suggests a methodological gap and an area yet to be fully explored where both knowledge of the world and interactive engagement with the environment converge.

\begin{figure}
    \centering
    \includegraphics[width=1\linewidth]{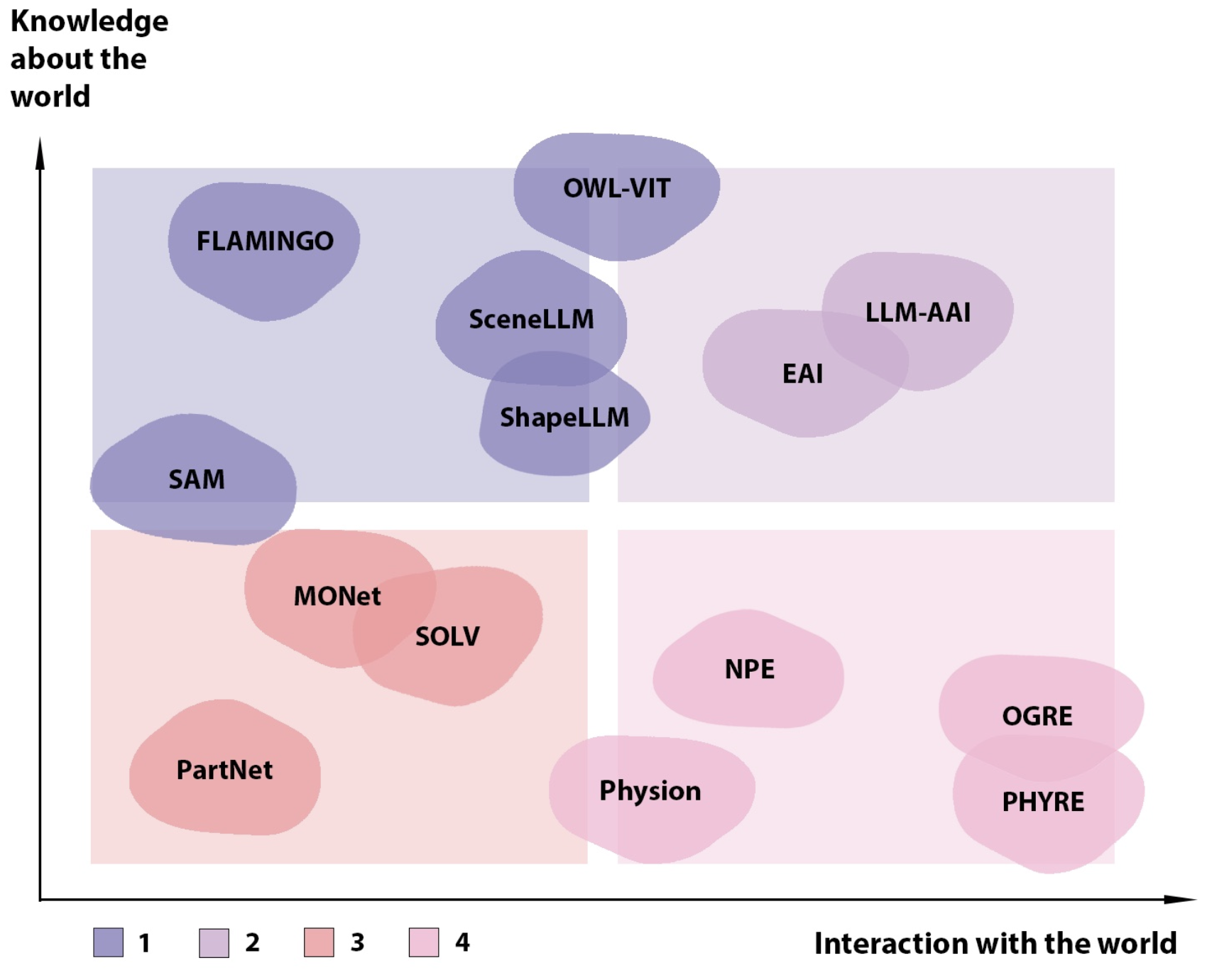}
    \caption{The four conceptual quadrants of paradigms of objecthood in AI and the example paradigms discussed in this section. Top-left quadrant (1): No Interaction, High Knowledge about the World Object Understanding. Top-right quadrant (2): Combining Knowledge of the World with Interaction with the World. Bottom-left quadrant (3): No Interaction, Low Knowledge about the World Object Understanding.  Bottom-right quadrant (4): High Interaction, No Knowledge about the World Object Understanding.}
    \label{fig:enter-label}
\end{figure}

In the subsequent sections, examples and benchmarks of each of these approaches will be covered in detail, including their methodologies, evaluation and the results and theoretical insights they provide for the field of AI object understanding. We start each section with the criteria that define the paradigm. 

\subsection{No Interaction, Low Knowledge about the World Object Understanding}

Paradigm Criteria:
\begin{itemize}
    \item Method: Relies on geometrical or visual segmentation, such as computer vision methodologies for object recognition, segmentation, or detection. 
    \item Objecthood Conceptualisation: Visual pattern detection, i.e. segmenting and extracting geometric object features such as edges, textures, and shapes from images or video data. 
    \item Data: Must be visual, such as still images or video frames, with an emphasis on pixel-level processing.
    \item Evaluation: How well the method segments objects within images based on geometric characteristics, i.e. Intersection over Union (IoU) or pixel-level accuracy.
\end{itemize}

Segmentation-based approaches to object understanding in AI are mainly grounded in deep learning methodologies of computer vision, where convolutional neural networks (CNNs) have been the most popular models for more than a decade. These models are widely used for image classification, object detection, and semantic segmentation, where they have demonstrated significant effectiveness in extracting spatial hierarchies and identifying features within visual data \cite{Chai2021}. Recent approaches also extend the method to video object segmentation \cite{gao2023deep}, where CNN models are a common approach for frame-by-frame object detection, allowing the segmentation of moving objects across time. The extension to video segmentation is important, as it facilitates object recognition in complex, temporally-variant environments, with implications for applications in e.g. autonomous driving, surveillance, and augmented reality, where real-time object understanding is required. The following are the most prominent benchmarks to date related to segmentation-based object understanding in AI.

MONet (Multi-Object Network) \cite{Burgess2019} presents a paradigm focusing on unsupervised scene decomposition and representation using a variational autoencoder (VAE) paired with a recurrent attention network to segment and reconstruct images. The segmentation process is unsupervised, with the model learning to identify coherent and distinguishable objects based purely on visual input data. MONet is evaluated primarily on ability to segment and reconstruct images, and focus on generalisation to complex visual scenes. The metrics used for evaluation include pixel-wise reconstruction accuracy and qualitative segmentation performance. In testing, MONet is evaluated on the Objects Room dataset, and achieves accurate segmentation of individual objects, walls, floors, and backgrounds in scenes containing 1 to 3 objects, as well as tested with more complex scenes, including those with up to 6 objects, the model successfully generalises and maintains segmentation accuracy. The model's performance on the CLEVR dataset (more visually complex 3D rendered objects) and the Multi-dSprites dataset (2D sprite-based images) also resulted in robust segmentation and high quality reconstructions. 

A related benchmark is PartNet \cite{Mo2019}, a large-scale dataset which provides detailed annotations of 3D models, building upon ShapeNet \cite{Chang2015}. The dataset is structured to allow multi-level segmentation, ranging from coarse object parts to fine-grained details, supporting semantic, hierarchical, and instance-level segmentation. PartNet benchmarks the performance of several 3D deep learning algorithms across three tasks: fine-grained semantic segmentation, hierarchical semantic segmentation, and instance segmentation of the dataset. The models tested include PointNet, PointNet++, SpiderCNN, and PointCNN for semantic segmentation, with different hierarchical segmentation approaches, including bottom-up, top-down, and ensemble methods, used for comparison. The metrics used for evaluation are primarily mean Intersection-over-Union (mIoU) for segmentation tasks and mean Average Precision (mAP) for instance segmentation. In fine-grained semantic segmentation, the best-performing models, such as PointNet++, achieved mIoU mean scores up to 65.5\%. In hierarchical segmentation, the ensemble method performed slightly better than bottom-up and top-down approaches, achieving an average mIoU of 51.7\% across categories. For instance segmentation, PartNet's benchmark reported mAP scores averaging 54.4\%.

As an example of a video-based approach, SOLV \cite{Aydemir2023} is a self-supervised method for multi-object segmentation in real-world video sequences. The model is trained using a masked autoencoder paradigm, where only a portion of the input observations is used to reconstruct dense visual features and is evaluated on both synthetic and real-world datasets. Performance is measured using the foreground adjusted Rand index (FG-ARI) for synthetic data and the mIoU metric for real-world datasets. To ensure consistency between frames, Hungarian Matching is applied to compare predicted segmentation masks with ground-truth annotations over time. 

Segmentation-based approaches to object understanding reveal several underlying assumptions about objecthood and its representation in AI models. Each benchmark method presumes that objecthood can be captured by visual segmentation based on spatial, geometric, or compositional attributes detectable through pixel-level or 3D data. This reflects an assumption that objecthood can be represented by visual features and relationships in the spatial hierarchy, overlooking potentially complex conceptual or contextual characteristics that contribute to our understanding of objects. MONet and PartNet largely focus on identifying and differentiating individual objects or object parts in still images or static 3D models, emphasising visual coherence and segmentation fidelity as measures of success. This approach is expanded in video segmentation methods like SOLV, which prioritises temporal consistency and multi-object tracking, acknowledging that object understanding in dynamic environments requires models to maintain coherence across frames. 

\subsection{No Interaction, High Knowledge about the World Object Understanding}

Paradigm Criteria:
\begin{itemize}
    \item Method: Relies on generative AI and prompting methodologies, i.e. employ zero- or few-shot learning techniques.
    \item Objecthood Conceptualisation: Identifying object relationships and segmenting objects based on learned contextual representations rather than solely geometric features.
    \item Data:  Input data can be either text, images or sequences (e.g., video) broken down into patches or tokens.
    \item Evaluation: Model generalisation across objecthood related tasks and domains (e.g., zero-shot or few-shot learning capabilities).
\end{itemize}

Methods that use a high knowledge about the world paired with low, to no interactions are for example generative AI approaches, such as LLM which can be employed for object understanding within image and video scene analysis by using language-based contextual comprehension capabilities \cite{kirillov2023segment, zhou2024survey}. To exemplify, LLMs may integrate semantic understanding of objects, actions, and events by aligning textual prompts with visual features extracted from video frames, which allows for associating objects with descriptive terms, enabling tasks such as object detection and classification based on contextual cues \cite{zhou2024survey}. Specifically, LLMs can guide video segmentation by generating textual descriptions or prompts that highlight desired objects or regions of interest, using their ability to model temporal relationships and semantic context across frames. Techniques such as transformer-based architectures have shown promise in aligning video scene dynamics with descriptive terms, improving the accuracy and coherence of tasks like action recognition and event detection \cite{zhou2024survey}. 

The Segment Anything benchmark \cite{kirillov2023segment} introduces an approach to object understanding through a foundation model designed for prompt-based, zero-shot segmentation. The approach uses a "promptable segmentation task," which allows the model to respond to any prompt – such as points, boxes, or text – that specifies what part of an image to segment. This capability is achieved by the Segment Anything Model (SAM), which uses a Vision Transformer (ViT) encoder and a lightweight prompt and mask decoder trained on a large-scale dataset. SAM's design aims to generalise across diverse segmentation tasks and data distributions without requiring specific task training. SAM’s performance is evaluated using a suite of 23 datasets with a range of segmentation challenges. Evaluation metrics include mIoU and human ratings, with single-point prompts as a central focus to test SAM's ambiguity handling and zero-shot adaptability. Additionally, SAM’s segmentation capability is compared against established interactive segmentation tools, and it undergoes zero-shot transfer tests in tasks like edge detection and object proposal generation. SAM outperforms baseline methods in zero-shot single-point segmentation on 16 out of 23 datasets, with mIoU improvements of up to 47 IoU points. In human evaluations, SAM’s segmentation quality received ratings between 7 and 9 on a 1–10 scale. 

Another system is Flamingo \cite{Alayrac2022}, a Visual Language Model (VLM) designed for few-shot learning in multimodal tasks. Flamingo combines pre-trained vision-only and language-only models through a new architecture that processes text together with images or videos. Flamingo is trained on a range of different multimodal datasets with text-image and text-video pairs. The model processes up to 32 visual-text pairs during evaluation, even though trained on sequences of only 5 pairs, showing some capacity for in-context learning. Performance is evaluated across 16 multimodal benchmarks, covering tasks from visual question answering to image captioning. Few-shot learning is assessed by prompting the model with task-specific examples without task-specific fine-tuning. Benchmarks include standard datasets like VQAv2 and VizWiz for question-answering and captioning tasks. Results establish significant results in few-shot learning across several tasks, including 6 tasks where it outperforms fine-tuned models using only 32 examples. The model achieves competitive results on multiple tasks, demonstrating adaptability and efficiency in vision-language alignment. 

SceneLLM \cite{fu2024scene} and ShapeLLM \cite{Qi2025} are systems that integrate language processing with 3D visual understanding. Both models utilise 3D spatial information to enhance language comprehension and generate contextually relevant responses to spatial tasks. Scene-LLM focuses on indoor environments, combining egocentric and scene-level information for tasks such as dense captioning, object interaction, and scene-based question answering (VQA). ShapeLLM, by contrast, centres on detailed object understanding using 3D point clouds and specialises in embodied interaction tasks. The key conceptual advance is the ability to mix 3D spatial representations with language prompts, enabling zero-shot learning and high adaptability across various multimodal tasks. 

Finally, an approach combines a VLM and an LLM to enhance robotic perception for handover tasks \cite{huang2024combining}. This system uses OWL-ViT for object detection and Llama3 for semantic understanding, enabling a robot to identify optimal grasping parts of an object in a zero-shot manner. The model aims to facilitate smoother handovers by integrating semantic knowledge about objects and their functional grasping parts, with implications for improving human-robot interaction. The method involves three stages: detection, semantic understanding, and segmentation. First, OWL-ViT detects and localises the object. Then, Llama3 interprets the object’s structure to identify the best parts for grasping. Finally, Grounded-SAM applies instance segmentation on the identified grasping regions to facilitate precise handover actions. The approach is evaluated across nine common objects (e.g., scissors, hammer, screwdriver) in a mixed simulation and real-world setting. Performance metrics include object detection accuracy and qualitative assessments of grasp part identification. The method’s robustness is tested in zero-shot conditions, assessing its adaptability to diverse objects without specific prior training.

Theoretically, generative AI and few-shot prompting methods approach object understanding with an emphasis on contextual, relational, and functional knowledge, rather than purely geometric features. This enables models to segment, identify, and relate objects based on prompts or descriptions, rather than strict spatial boundaries. These approaches suggest that objecthood involves understanding relationships and functionalities in context, which allows them to generalise across tasks and domains, as seen by zero- and few-shot learning abilities. However, this assumption also implies a core limitation: whilst capable of understanding object roles and relations based on large-scale pre-trained knowledge, these methods have limited interaction with objects and environments. Without interaction, their object understanding remains relatively passive, constrained by pre-existing contextual information rather than active, experiential engagement. This lack of direct interaction could restrict their utility in tasks where physical engagement or adaptation to dynamic environments is crucial.

\subsection{High Interaction, No Knowledge about the World Object Understanding}

Paradigm Criteria:
\begin{itemize}
    \item Method: DRL or simulation-based approaches, where agents interact with objects in virtual or game-engine environments to learn object properties and behaviours.
    \item Objecthood Conceptualisation: Revolves around dynamic and embodied object interaction, grounded in intuitive physics and ability to simulate real-world dynamics in virtual spaces.
    \item Data:  Primarily simulation-based task data which may not always involve pixel-level visual data but instead object-state representations in the simulation.
    \item Evaluation: Task success and performance, task completion rates, reward maximisation and task generalisation.
\end{itemize}

Game engines and dynamic simulation platforms provide foundational environments for AI to learn intuitive physics and object interactions through direct interaction and observation. This approach, seen in environment benchmarks like OGRE and PHYRE, enables AI agents to engage in tasks that require reasoning based on real-world physics principles—such as gravity, collision, and friction \cite{Allen2020, Bakhtin2019}. Unlike passive image-based methods, game-engine-based models involve DRL and simulation-based frameworks, where agents interact with objects to learn properties. Benchmarks in this domain evaluate models to generalise across novel objects and situations. OGRE \cite{Allen2020}, for example, is a physical reasoning environment designed to test an agent’s ability to generalise object-related knowledge to new contexts and object types, whilst PHYRE \cite{Bakhtin2019} emphasises sample efficiency, requiring agents to achieve goals (e.g., making objects contact each other) with minimal attempts.. These benchmarks rely on task success rates, completion metrics, and reward maximisation as evaluation criteria, to ensure that models develop robust, adaptable, and efficient physical reasoning capabilities for real-world applications in robotics, autonomous systems, and embodied AI.

\cite{Chang2016} introduce the Neural Physics Engine (NPE), a neural network framework designed to simulate intuitive physics by factoring a scene into object-based representations. Inspired by compositional reasoning in physics engines, NPE models object dynamics through pairwise interactions, allowing for robust generalisation across variable object counts and scene configurations. This approach enables NPE to simulate physical interactions with minimal reliance on scene-specific retraining. The NPE uses a differentiable model that applies a neighbourhood mask to select relevant "context" objects for each focus object. Pairwise interactions are encoded and summed to represent the net effect on the focus object, allowing the model to predict future states (such as velocity) based on causal interactions. The NPE is tested on three main tasks: predicting motion in fixed-object scenes, generalising to new scenes with different object counts, and inferring latent properties such as object mass. Key evaluation metrics include cosine similarity and velocity mean squared error (MSE) between the predicted and ground truth trajectories. In generalisation tasks, NPE maintains stability, handling up to eight interacting objects with minimal error accumulation.

Physion \cite{Bear2022} is a dataset and benchmark designed to assess physical reasoning abilities. The benchmark focuses on evaluating predictive physical understanding, requiring models to predict interactions between objects (e.g., collisions, stability, containment) within diverse, visually realistic scenarios. Physion utilises the ThreeDWorld (TDW) simulator to generate a range of physical interactions across eight scenarios: Dominoes, Support, Collide, Contain, Drop, Link, Roll, and Drape. Each scenario includes a task setup in which an “agent” object’s interaction with a “patient” object forms the basis for prediction. Models are trained on visually observed inputs and assessed on their ability to predict outcomes, such as whether objects will come into contact. The benchmark involves models from various architectures, such as vision-based CNNs, RNN-based dynamics predictors, and particle-based models like Graph Neural Networks (GNNs). Evaluation metrics include prediction accuracy, Pearson correlation to average human responses, and Cohen’s kappa to measure agreement with human judgement. Models are tested in three training settings to assess their generalisation across different physical scenarios. Physion results indicate that particle-based models, which have direct access to simulated physical states, outperform purely vision-based models. Graph Neural Network models, in particular, approach human-level performance on tasks such as object stability and collision.

Game engines, such as Unity or Unreal Engine, are software frameworks designed for creating and rendering virtual environments, often used in video game development but increasingly used for AI research \cite{Shao2019}. These platforms allow for highly customizable environments with realistic physics, lighting, and object interactions, making them suitable for testing and evaluating AI agents. Older platforms like Project Malmo, built on Minecraft, have also been popular for AI research \cite{johnson2016malmo}. Malmo provides a simpler, block-based environment where agents can learn and test capabilities like navigation, object recognition, and task execution. Any game engine (such as Unity) can be combined with a reinforcement learning interface to create testbeds for evaluating AI agents \cite{Burnell2022, crosby2019animal} re specific object understanding capabilities, such as object permanence \cite{Voudouris2024} and affordances \cite{Rutar2024}. For instance, the Animal-AI Environment, a virtual lab for cognitive tests on AI agents, includes tasks modelled on comparative cognition research, providing a systematic way to evaluate agent's understanding of object permanence \cite{Voudouris2022}. This setup allows agents to interact with objects and simulate real-world physical properties. One specific aspect of object understanding that has been investigated with the help of game engines is affordances, which is seen as a foundational cognitive capability for AI agents to understand, master and robustly evaluate \cite{Rutar2024}. The embodied interaction approach to affordance learning in AI agents is promising, as it allows for understanding affordances as a property of the objects themselves, the agents’ own properties and the environment. For instance, \cite{Nagarajan2020} introduced "interaction-exploration" agents that construct their own affordance models by actively engaging with and exploring their environment. Similarly, \cite{Liao2022} evaluated a specialised reinforcement learning model, implemented through a virtual robotic arm, to test its capacity for generalising learned affordance knowledge to unfamiliar objects.

These approaches to object understanding provide a theoretical approach to objecthood by emphasising direct interaction, intuitive physics, and dynamic responses within virtual environments. These benchmarks assume that engaging with objects in a simulated setting enables AI to develop an embodied understanding of physical properties, such as collision and stability, and to generalise across different object dynamics. Whilst effective at learning through simulated physics, these models are limited by their restricted semantic knowledge about the world within the virtual environment, often lacking contextual or functional understanding beyond physical interactions. This limitation constrains their application in tasks that require semantic nuance or contextual reasoning, as they cannot capture the fundamental meanings or uses of objects in real-world contexts. Thus, whilst useful for embodied interactions and dynamic tasks, these models still fall short of capturing a comprehensive notion of objecthood that includes semantic and situational understanding.

\subsection{Combining Knowledge of the World with Interaction with the World}
Paradigm Criteria:
\begin{itemize}
    \item Method: Combination of DRL/simulation-based approaches and generative AI and prompting methodologies, i.e. transformer models.
    \item Objecthood Conceptualisation: Combines dynamic and embodied object interaction with  learned contextual and semantic representations.
    \item Data:  Combination of pixel-level visual data but instead object-state representations in interaction and text, images or sequences (e.g., video) broken down into patches or tokens for prompting.
    \item Evaluation: Task success and performance, task completion rates, reward maximisation as well as generalisation across various objecthood related tasks and domains.
\end{itemize}

The Top-right quadrant of the framework remains under-explored, and is representing a space where both significant world knowledge and active, embodied interaction with the environment may be combined. Whilst only a few studies exist in this domain \cite{Li2024, Mecattaf2024} offers a promising initial attempt by combining game engine interaction with LLMs. This paper introduces a framework called LLM-AAI, which integrates LLMs with the Animal-AI (AAI) 3D simulation environment \cite{crosby2020animal}. The AAI environment mimics laboratory setups used in cognitive science to study animal intelligence and physical common-sense reasoning and this approach allows LLMs to directly control an agent within this 3D simulation, assessing how well these models can apply their internal knowledge to interact with physical objects and environments. The LLM-AAI framework provides a multi-modal interface where LLMs receive text and image inputs from the environment and return action plans in a simplified scripting language. The LLM-AAI framework is specifically designed to evaluate how well LLMs can reason about physical interactions and translate this reasoning into real-time actions. The study evaluated three state-of-the-art multi-modal LLMs—Claude 3.5 Sonnet, GPT-4o, and Gemini 1.5 Pro—on a subset of 40 tasks from the AAI Testbed. These tasks were designed to replicate cognitive science experiments and were previously completed by children. The performance of the LLMs was compared against the top 10 competition agents from the 2019 Animal-AI Olympics \cite{crosby2019animal} and human children aged 6-10. The results indicated that whilst LLMs could complete simpler tasks, their performance sharply declined as task complexity increased. This paper serves as an important early exploration of how LLMs might be "embodied" within interactive environments, which is an important step towards filling the gap represented by the top-right quadrant in Fig. 1. Although the results show that LLMs are not yet competitive with human children in complex physical reasoning tasks, the approach offers a robust foundation for further experimentation. 

Another paradigm example is Embodied Agent Interface (EAI) \cite{Li2024} which explores this space by introducing a systematic framework to evaluate how well LLMs can perform embodied decision-making tasks. The EAI framework unifies decision-making tasks across two simulation environments, VirtualHome and BEHAVIOR, enabling a standardised approach to assess LLMs in goal interpretation, subgoal decomposition, action sequencing, and transition modelling. The evaluation focused on 18 LLMs, including GPT-4o and Claude-3.5 Sonnet, and highlighted both strengths and limitations. Whilst models performed well in simpler tasks, performance declined significantly in complex environments, particularly where relational reasoning and spatial goal satisfaction were required. Errors included missing preconditions, additional unnecessary actions, and trajectory feasibility issues. Despite these challenges, EAI marks a critical step toward understanding how LLMs can interact dynamically within complex, embodied environments, addressing key gaps in the top-right quadrant and laying groundwork for future research. Future work could refine the control schemes, improve the sensory feedback for LLMs, and explore more sophisticated multi-modal training strategies to contribute towards object understanding that includes both knowledge about the world and interaction with the world.

See Table 1 below for the overview of the four approaches presented above and with them associated paradigms, methods, objecthood conceptualisations, data and evaluation metrics. 

\small
\renewcommand{\arraystretch}{1.5}
\setlength{\tabcolsep}{5pt}
\begin{longtable}{|p{2cm}|p{2cm}|p{2cm}|p{2cm}|p{2cm}|p{2cm}|p{2cm}|}
\caption{Overview of AI paradigms for object understanding, categorised by their interaction level with and knowledge about the world. For each approach, the table details the specific methods used, how objecthood is conceptualised, the types of data utilised, and evaluation metrics employed within each paradigm.} \\
 
\hline
\textbf{Paradigm} & \textbf{Approach} & \textbf{Method} & \textbf{Objecthood Conceptualization} & \textbf{Data} & \textbf{Evaluation} \\ \hline
\endfirsthead

\hline
\textbf{Paradigm} & \textbf{Approach} & \textbf{Method} & \textbf{Objecthood Conceptualization} & \textbf{Data} & \textbf{Evaluation} \\ \hline
\endhead

\hline
\endfoot

\hline
\endlastfoot

\multirow{3}{=}{\textbf{No Interaction, Low Knowledge about the World}} 
& MONet & Attention Network + Component VAE & Spatial visual recomposition and representation of 3D scenes into semantically meaningful components i.e. objects & Objects Room dataset: rendered 3D scene images & Pixel-wise reconstruction accuracy, qualitative segmentation performance \\ 
& PartNet & PointNet, PointNet++, SpiderCNN, PointCNN & 3D model semantic segmentation, hierarchical semantic segmentation, and instance segmentation & 3D model object dataset based on ShapeNet & mIoU, mAP \\ 
& Aydemir et al. & Transformer variant: masked autoencoder & Multi-object segmentation in real-world video sequences. & Two real-world video datasets: MOVi-E synthetic dataset + YouTube-VIS 2019 & FG-ARI, mIoU \\ \hline

\multirow{3}{=}{\textbf{No Interaction, High Knowledge About the World}} 
& Segment Anything & Vision Transformer (ViT) encoder + prompt encoder + mask decoder & Prompt-based (points, boxes, or text) semantic zero-shot specification of visual segmentation & 23 datasets with segmentation challenges. & mIoU, Human ratings, Zero-shot adaptability and zero-shot transfer tests \\ 
& Flamingo & Visual Language Model (VLM): vision encoder + Perceiver-based resampler + gated cross-attention layers & Few-shot learning in multimodal tasks combining pre-trained vision-only and language-only models through processing text together with images or videos. & Multimodal datasets with text-image and text-video pairs. & Visual question answering, Image captioning \\
& Scene-LLM & LLM pre-trained with paired 3D frame-language data and fine-tuned with instruction-following data & Using 3D spatial information to enhance language comprehension and generate contextually relevant responses to spatial tasks. & Visual 3D indoor environments, combining egocentric and scene-level information & BLEU, Exact Match \\ \hline

\multirow{2}{=}{\textbf{High Interaction, No Knowledge about the World}} 
& OGRE & Random agent (RAND), Object-Oriented Random Agent (OORAND), Deep Q-network (DQN) & Agent interaction in physical reasoning and generalization simulated environment & Test suite of tasks i.e. puzzles with goal states. & Success rate \\ 
& PHYRE & Random agent (RAND), Non-parametric agent (MEM), Non-parametric agent with online learning (MEM-O), Deep Q-network (DQN), Deep Q-network with online learning (DQN-O) & Agent interaction in simulated intuitive physics 2D environment for visual reasoning. & Test suite of tasks i.e. classical mechanics puzzles consisting of initial world state + goal. & Success rate, AUCCESS \\ \hline

\multirow{2}{=}{\textbf{Combining Knowledge of the World with Interaction with the World}} 
& LLM-AAI & Claude 3.5 Sonnet, GPT-4o, and Gemini 1.5 Pro + Animal-AI (AAI) 3D simulation environment & LLM controlled agent within 3D simulation applying internal knowledge to interact with physical objects and environments. & Set of 40 tasks from the AAI Testbed designed to replicate cognitive science experiments. & Success rate \\ 
& EAI & 18 LLMs, including i.e. GPT-4o and Claude-3.5 + VirtualHome and BEHAVIOR simulation environments & LLM embodied decision-making tasks in simulated environments, i.e. object-centric modeling: states as relational features among entities in the environment. & Tasks i.e. language-based, object-centric problem representations with objects, states, actions. & Goal interpretation (ground truth comparison), trajectory feasibility and goal satisfaction scores, partial goal satisfaction evaluation, logic matching score, planning success \\ \hline

\end{longtable}

\subsection{Overall Analysis}
Based on the paradigms presented above, we come to two conclusions. First, the AI systems presented in the Top-left quadrant (1), Bottom-left quadrant (3) and the Bottom-right quadrant (4) possess at least some level of each core capability related to object understanding. In these cases, individual accuracy scores and evaluation metrics could undoubtedly be improved further, particularly in terms of robustness. However, the broader issue lies in the narrow behavioural focus of artificial systems, which are often designed to excel at highly specific tasks targeting isolated aspects of object understanding. This limited scope hinders the development of systems capable of a more comprehensive and integrated understanding of objects. This leads us to the second observation, which is that the artificial systems presented in the Top-left quadrant (1), Bottom-left quadrant (3) and the Bottom-right quadrant (4) do in fact possess a high degree of functional integration across the core capabilities related to object understanding. We suggest that artificial systems designed that way should pave the way forward. Let us expand on these two observations.

\textbf{Narrow focus on individual capabilities: Top-left quadrant (1), Bottom-left quadrant (3) and the Bottom-right quadrant (4)}

Why exactly is a narrow focus problematic in the context of developing world models with a robust sense of objecthood?
The core problem lies in the interdependent nature of the core capabilities related to object understanding. As outlined in the cognitive science section, these capabilities – perceptual grouping (c1), behavioural grouping (c2), and affordance identification (c3) – are inherently intertwined. So, if one capability, such as perceptual grouping (c1), relies on the possession of two other capabilities, behavioural grouping (c2) and affordance identification (c3), then all three must be effectively modelled in artificial systems. However, most often, this is not the case. Modellers may know how to design agents with c1 but not also c2 and c3, or vice versa. As a result, when trying to simulate c1, they often rely on pre-computed outputs that c2 and c3 would provide if those capabilities were present, thereby circumventing the need for fully integrated modelling.

The Bottom-left quadrant (3) exemplifies this limitation through its emphasis on perceptual grouping (c1). Whilst agents in this quadrant are designed to segment objects from their environment, they often do so without incorporating the dynamic interactions that behavioural grouping (c2) requires or the functional insights afforded by affordance identification (c3). For instance, the MONet benchmark (Burgess et al., 2019) demonstrates unsupervised segmentation of complex scenes, excelling at perceptual grouping but failing to model behavioural dynamics or affordance reasoning.
The Bottom-right quadrant (4), on the other hand, emphasises behavioural grouping (c2), focusing on how objects interact in dynamic contexts. However, the reliance on pre-segmented inputs for behavioural grouping means that these agents are not required to autonomously segment objects or discern their identities (c1), and they often fail to link these dynamics to actionable affordances (c3).The Physion benchmark (Bear et al., 2022) highlights this focus, evaluating agents on physical reasoning tasks like object collisions or stability but assuming that segmentation (c1) is already solved. Note though that as part of the emphasis on the interaction in dynamic contexts, AI systems are being developed that are particularly adept at recognising and using affordances \cite{Liao2022, Nagarajan2020}. Thus, combining AI systems that focus on behavioural grouping (c2) and affordances understanding (c3) in the context of skilful interaction with the environment should be the priority.

The Upper-left quadrant (1) contains the models with the ability to discern what objects can do within specific contexts. These models excel at associating objects with their potential uses by leveraging contextual and relational knowledge, capturing limited affordance understanding capabilities (c3). However, they often lack perceptual grouping (c1), as they rely on pre-segmented inputs, and they do not model dynamic interactions, which are essential for behavioural grouping (c2). Flamingo \cite{Alayrac2022}, for example, demonstrates the ability to align textual and visual data, excelling at identifying object affordances based on descriptive cues. However, its reasoning remains static, as it lacks integration with perceptual segmentation (c1) or dynamic behaviour modelling (c2). SceneLLM \cite{fu2024scene}, with its integration of 3D spatial representations, enhances affordance reasoning by contextualising object interactions in indoor scenes. Yet, like Flamingo, it relies on pre-defined inputs and does not autonomously predict object behaviours (c2) or perform segmentation (c1).

This piecemeal approach underscores a major limitation: whilst each quadrant addresses one or more individual capabilities, none achieves the functional integration necessary for robust object understanding. Without integrating these capabilities, agents remain confined to narrowly defined tasks, unable to form a cohesive and flexible understanding of objects.

\textbf{Broader focus on interrelated capabilities: Top-right quadrant (2)
}

The LLM-AAI framework \cite{Mecattaf2024} presented in the Top-right quadrant (2), however, represents a significant step toward overcoming this limitation by explicitly combining multiple modalities  – textual, visual, and interactive inputs – with real-time agent interaction. The integration achieved by this framework bridges the gap between isolated tasks and cohesive, multi-faceted object understanding. For example, when an LLM-driven agent interacts with the 3D simulation environment, it must use visual inputs to identify and segment objects (c1), contextual knowledge to predict object behaviours (c2), and reasoning capabilities to understand affordances (c3). These capabilities are not treated as separate modules but as part of a unified pipeline, where the outputs of one process inform and enhance the others.

By designing tasks that mimic cognitive science experiments and involve elements like stacking, containment, and navigation, the LLM-AAI framework effectively illustrates the interplay between capabilities \cite{Mecattaf2024}. For instance, successful completion of these tasks requires the agent to integrate affordance reasoning with perceptual and behavioural insights, demonstrating a level of functional interdependence that is absent in systems constrained to individual benchmarks. This dynamic interaction is crucial because it allows the agent to adapt its reasoning and behaviour to new scenarios, moving beyond static, pre-solved inputs like ground truth segmentations.

Despite these promising results, the foundation models used in the Top-right quadrant (2) faces some limitations. For example, with the current technology they can only be updated sparsely (through finetuning) and the embodiment is posterior to the original training that led to the knowledge about the objects. Additionally, for this approach to truly resemble human object understanding, the right way to interrelate objecthood capabilities will need to be found. These limitations prevent them from fully committing to the goals of the integrated approach. In short, the LLM-AAI framework \cite{Mecattaf2024} underscores the broader point that genuine object understanding cannot emerge from piecemeal efforts but requires a cohesive architecture where capabilities co-evolve.

\section{Conclusion and Way Forward}

The goal of this paper was twofold. First, to provide an overview of the ways in which object understanding has been studied in cognitive science, thus helping bring disparate research communities together and highlighting their unique contributions to object understanding. We presented three distinct theoretical frameworks that tackle the question of object understanding where each framework assumed a different core capability which fulfils a particular functional role of perception – but importantly, for a robust sense of objecthood an integration across these distinct functional goals is necessary. This is by no means supposed to be an exhaustive list of core capabilities that underlie object understanding but rather an overview of some of the key capabilities that have been explored in cognitive science in this context. This part of our theoretical investigation also illustrates the importance of a robust sense of objecthood as a crucial component of world models, enabling prediction and interaction with the world. Core objecthood capabilities are involved in the very initial scene construction and segmentation, but also allow us to make predictions about how the objects around us will behave and how to use them skillfully. These core objecthood capabilities further serve as precursor capabilities for many other complex capabilities, such as tool use, problem solving, and planning, that allow us to effortlessly exist in the physical world.

Our second aim was to evaluate the extent to which AI has been able to integrate findings from cognitive science about object understanding. We found that whilst AI systems show some progress in isolated object capabilities, they lack the functional integration observed in biological agents. The Bottom-left quadrant (3) emphasises perceptual grouping (c1), achieving success in tasks like segmentation but failing to incorporate dynamic behavioural predictions (c2) or affordance reasoning (c3). The Bottom-right quadrant (4) focuses on behavioural grouping, enabling agents to model object interactions (c2) in dynamic contexts and recognise and act upon affordances to some degree (c2), yet it relies on pre-segmented inputs and lacks the capacity for autonomous perceptual grouping (c1). Similarly, models in the Bottom-left quadrant (1) leverage contextual and relational knowledge, capturing some affordance understanding capabilities (c3), but they often do so using structured inputs, bypassing the need for agents to segment objects (c1) or understand their dynamic behaviour (c2).

A promising way forward is illustrated by approaches that combine existing world knowledge with real-time interaction, as seen in the LLM-AAI framework (Mecattaf et al., 2024) presented in the Top-right quadrant (2). This approach merges language models with dynamic 3D environments, enabling agents to process textual and visual information whilst interacting with objects. Such frameworks allow agents to link abstract knowledge with physical interactions, supporting a deeper, multi-modal understanding of objects that is both contextually adaptive and more functionally integrated. Nevertheless, even the foundation models used in this framework faces some important limitations; currently, they can only be updated sparsely (through finetuning) and the embodiment is posterior to the original training that led to the knowledge about the objects. This prevents them from fully fulfilling the goals of the integrated approach. Moving toward such frameworks like the LLM–AAI could work as a starting point for bridging the gap between isolated object capabilities and a more holistic, general-purpose AI, ultimately advancing models toward genuine object understanding in real-world scenarios.

Perhaps the narrow focus of artificial systems in the context of object understanding is partly the result of the highly disunified research on object understanding in cognitive science. Each theoretical framework emphasises a different core capability and a different functional goal of perception, and there is very little if any cross-pollination between these frameworks. It is our hope that bringing together these accounts as we have done here highlights the importance and unique contribution of each theoretical framework, as well as considerable areas of overlap and agreement. Such recognition has the potential to contribute to a more multifaceted research approach in AI that integrates across capabilities and goals, and can encourage a dialogue between disparate communities in cognitive science and AI, which might in turn lead to a more unifying framework for studying object understanding.

\section*{Declarations}

Funding: This work was supported by US DARPA [HR00112120007] (RECoG-AI). 
JHO was also funded by CIPROM/2022/6 (FASSLOW), 
the EC H2020-EU grant agreement No. 952215 (TAILOR), 
and Spanish grant PID2021-122830OB-C42 (SFERA) [MCIN/AEI/10.13039/501100011033].

\bibliography{sn-bibliography}% common bib file
%% if required, the content of .bbl file can be included here once bbl is generated
%%\input sn-article.bbl

\end{document}